\documentclass[10pt,twocolumn]{article} 
\usepackage{simpleConference}
\usepackage{times}
\usepackage{graphicx}
\usepackage{amssymb}
\usepackage{url,hyperref}
\usepackage{CJK}
\usepackage{makecell}
\usepackage{natbib}
\usepackage{multibib}
\setcitestyle{authoryear}
\usepackage{mdframed}
\usepackage{listings}
\usepackage{multirow}
\usepackage{verbatim}
\usepackage{makecell}
\usepackage{setspace}

\usepackage[T1]{fontenc}

\usepackage{lipsum}

\title{Exploring ChatGPT's Ability to Rank Content: A Preliminary Study on Consistency with Human Preferences}

\author{Yunjie Ji, Yan Gong, Yiping Peng, Chao Ni, Peiyan Sun, Dongyu Pan, Baochang Ma\thanks{Corresponding author.} , Xiangang Li \\
Beike Inc., Beijing, China  \\
\texttt{\{jiyunjie001, gongyan013, pengyiping001,nichao007, sunpeiyan001,}  \\
  \texttt{pandongyu002, mabaochang001, lixiangang002\}@ke.com}}
  
\begin{document}
\maketitle
\begin{abstract}
As a natural language assistant, ChatGPT is capable of performing various tasks, including but not limited to article generation, code completion, and data analysis.
Furthermore, ChatGPT has consistently demonstrated a remarkable level of accuracy and reliability in terms of content evaluation, exhibiting the capability of mimicking human preferences. 
To further explore ChatGPT's potential in this regard,  a study is conducted to assess its ability to rank content.
In order to do so,  a test set consisting of prompts is created, covering a wide range of use cases, and five models are utilized to generate corresponding responses. 
ChatGPT is then instructed to rank the responses generated by these models.
The results on the test set show that ChatGPT's ranking preferences are consistent with human to a certain extent.
This preliminary experimental finding implies that ChatGPT's zero-shot ranking capability could be used to reduce annotation pressure in a number of ranking tasks.
\end{abstract}

\section{Introduction}

Large Language Models (LLMs) \cite{LanguageModelsAre2020, OPTOpenPretrained2022, PaLMScalingLanguage2022, BLOOM176BParameterOpenAccess2022, GPTNeoX20BOpenSourceAutoregressive2022, TrainingComputeOptimalLarge2022} have made significant contributions to the development of natural language processing (NLP), especially in the performance of zero-shot and few-shot tasks.
Through instruction tuning \cite{FinetunedLanguageModels2021, MultitaskPromptedTraining2021, CrosslingualGeneralizationMultitask2022, ScalingInstructionFinetunedLanguage2022} and alignment training \cite{TrainingHelpfulHarmless2022, TrainingLanguageModels2022}, LLMs have become more than just general-purpose NLP task solvers \cite{HolisticEvaluationLanguage2022, ChatGPTGeneralPurposeNatural2023}, but also natural language assistants capable of performing various text-related tasks, such as writing articles, completing code \cite{EvaluatingLargeLanguage2021, CodeGenOpenLarge2023}, and analyzing structural data.

Recent studies have shown that ChatGPT is not only capable of generating high-quality content as requested by users, but also capable of providing effective evaluations of content \cite{LargeLanguageModels2023, ChatGPTGoodNLG2023}. 
By using carefully designed prompts, ChatGPT can provide reasonable evaluations of given content from different perspectives.  
Currently, most research has focused on using ChatGPT to evaluate a single sample, and its ability to rank multiple samples has not been well explored.

In this paper, we explore this question and construct a set of prompts covering typical NLP tasks, brainstorming, article generation, and other open-ended tasks. 
For each prompt, five models are adopted to generate responses.
By comparing ChatGPT's ranking results to human preference annotations, we find that ChatGPT's ranking preferences had a certain level of consistency with humans. 
Based on this finding, ChatGPT's zero-shot ranking capabilities can be leveraged  to obtain more annotation data and improve the performance of the model on ranking tasks. 
For example, in RLHF \cite{TrainingLanguageModels2022, TrainingHelpfulHarmless2022}, ChatGPT can be used to construct a set of initial preference annotation data for bootstrapping the reward model.

\section{Related Work}

\begin{figure*}[t!]
	\centering
	\includegraphics[scale=0.55]{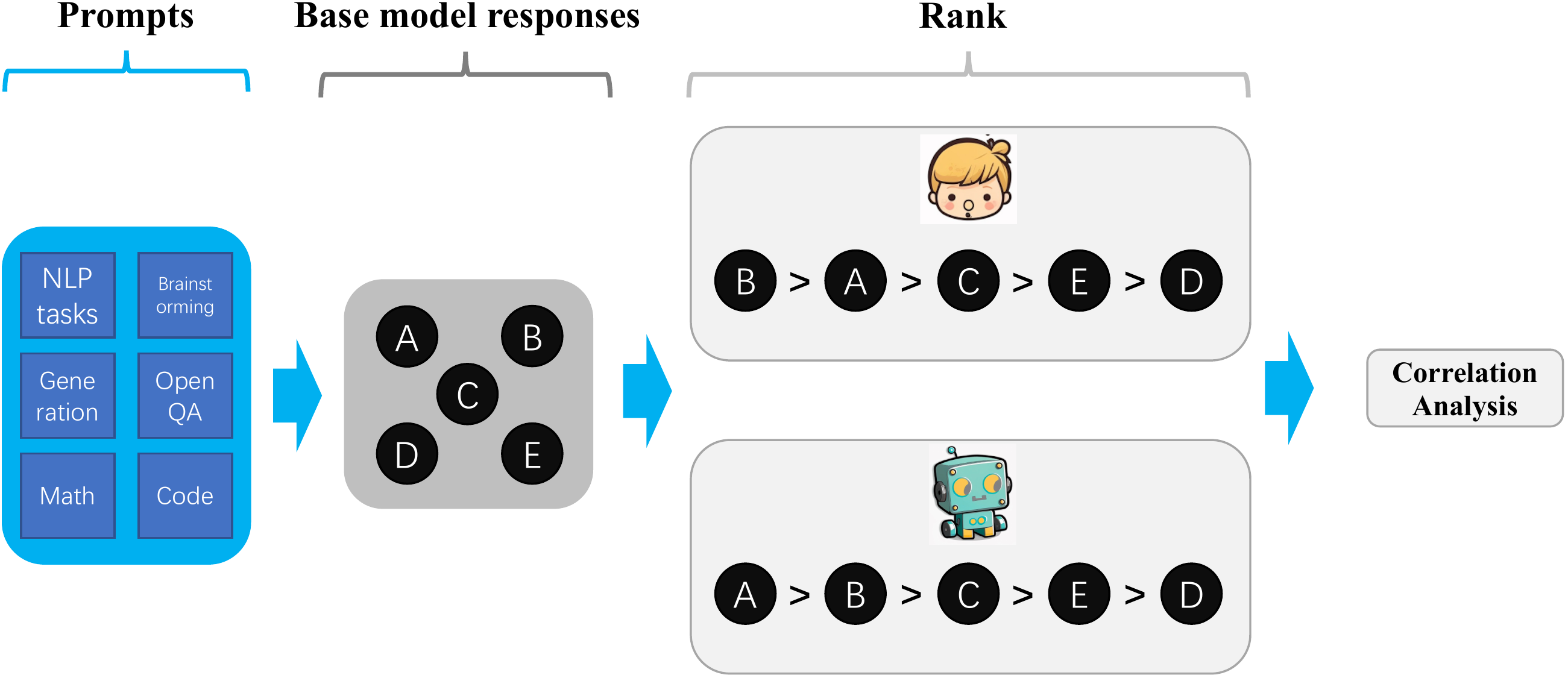}
	\caption{The illustration of our evaluation process.}
\label{procedure}
\end{figure*}

The models based on transformer \cite{vaswani2017attention, devlin2018bert, lan2019albert, yang2019xlnet, dong2019unified, clark2020electra, raffel2020exploring} have greatly promoted the development of NLP, and recently the generative large language models\cite{LanguageModelsAre2020, OPTOpenPretrained2022, PaLMScalingLanguage2022, GPTNeoX20BOpenSourceAutoregressive2022, TrainingComputeOptimalLarge2022, glaese2022improving, srivastava2022beyond} have made further progress.
One notable example is the GPT (Generative Pre-trained Transformer) family of models, ChatGPT has further improved its ability to understand and follow human instructions by RLHF \cite{TrainingLanguageModels2022, ConstitutionalAIHarmlessness2022, FineTuningLanguageModels2020, LearningSummarizeHuman2022, RedTeamingLanguage, WebGPTBrowserassistedQuestionanswering2022, PretrainingLanguageModels2023}. 
This has transformed ChatGPT  from a simple NLP task solver to a comprehensive natural language assistant capable of tasks such as writing articles, analyzing structural data.

Using pre-trained language models as the evaluation metric (e.g., BERTScore \cite{zhang2019bertscore}, MoverScore\cite{zhao2019moverscore} , COMET\cite{rei2020comet} , BLEURT\cite{sellam2020bleurt} , MAUVE\cite{pillutla2022mauve} and BARTScore\cite{yuan2021bartscore} receives increasing attention.
Considering ChatGPT's impressive capability as an intelligent conversational language model, there are works studying ChatGPT from various perspectives. 
\cite{ChatAugLeveragingChatGPT2023} uses ChatGPT to generate augmented data to further improve the performance of the model.
\cite{ChatGPTBeginningEnd2023} found that ChatGPT's zero-shot ability in genre classification tasks exceeded that of the multilingual XLM-RoBERTa language model, which was fine-tuned on training data. Similarly, \cite{tangdoes} utilized ChatGPT to generate high-quality synthetic data, reducing the time and effort required for data collection and labeling, while also addressing data privacy concerns and enhancing the ability to perform clinical text mining.
This demonstrates ChatGPT's strength as a data annotator. 
In addition to generating and annotating data, ChatGPT is also a great evaluator of texts. 
For example, \cite{LargeLanguageModels2023} used ChatGPT to evaluate translation model results and found that its accuracy reached SOTA in WMT22's Metrics shared task. 
\cite{ChatGPTGoodNLG2023} evaluated ChatGPT's performance on NLG tasks and found that it surpassced previously widely-used automatic NLG metrics, achieving the highest consistency with human-annotated results.

Our work is similar to \cite{LargeLanguageModels2023} and \cite{ChatGPTGoodNLG2023}, but we focus on ChatGPT's ability to rank the generated contents of different models. 

\begin{table*}[t!]
\caption{Statistics of test prompts.}
\small
\begin{center}
\begin{tabular}{c|c|c|c} 
\hline 
\textbf{Use case} & \textbf{\#Nums} & \textbf{Ave prompt length} & \textbf{Prompt example} \\
\hline   
NLP tasks & 600 & 64.79  & \makecell[l]{ \begin{CJK*}{UTF8}{gkai}我想知道下面两句话的意思是否相同。“花呗分期了会影响芝麻分吗”，“花 \end{CJK*} \\ \begin{CJK*}{UTF8}{gkai} 呗被限额度了会不会有什么影响”。选项：是的,不是 \end{CJK*} } \\
\hline
Brainstorming & 611 & 20.44  & \makecell[l]{\begin{CJK*}{UTF8}{gkai} 如何克服焦虑？ \end{CJK*} }\\
\hline
Generation & 605 & 37.14  & \makecell[l]{\begin{CJK*}{UTF8}{gkai} 以成功为主题，撰写一篇商业风格的人物访谈，探讨如何将某种难题或挑战 \end{CJK*} \\ \begin{CJK*}{UTF8}{gkai} 转化为商业机会。 \end{CJK*}}\\
\hline
Open QA & 584 & 15.68  & \makecell[l]{\begin{CJK*}{UTF8}{gkai} 鸡蛋的蛋黄中含有增强人脑记忆不可缺少的哪种物质？\end{CJK*}}\\
\hline
Math & 106 & 68.41  & \makecell[l]{\begin{CJK*}{UTF8}{gkai}红，黄，蓝气球共有62只，其中红气球的五分之三等于黄气球的三分之二，\end{CJK*}\\ \begin{CJK*}{UTF8}{gkai}蓝气球有24只，红气球和黄气球各有多少只？\end{CJK*}} \\ 
\hline
Code & 494 & 65.59  & \makecell[l]{\begin{CJK*}{UTF8}{gkai}请用Python语言完成如下题目解答。有四个数字：1、2、3、4，能组成多\end{CJK*}\\ \begin{CJK*}{UTF8}{gkai}少个互不相同且无重复数字的三位数？各是多少？\end{CJK*}} \\
\hline
\end{tabular}
\end{center}
\label{data_stat}
\small
\end{table*}

\section{Method}
In order to enable ChatGPT to accomplish ranking tasks,  two types of prompts are designed: one-by-one ranking prompt and one-for-all ranking prompt. 
The first prompt is as follows:
\begin{mdframed}
Please give a score to the answer, ranging from 1 to 5. If the response does not meet the requirements of the instruction, the score is 1. Only give a score, no additional explanation is needed. {``Instruction'': ``...'',``Response'': ``...''}
\end{mdframed}
 In this prompt, ``Instruction'' is from our test set and ``Response'' is generated by one of our  models.
 Therefore ChatGPT is actually required to assess the  model against following user instructions.
The second prompt is as follows:
\begin{mdframed}
Given a JSON file that contains an instruction and five potential responses, please rank the quality of these responses, and return the index of the ranked responses, no additional explanation is needed,  \{``Instruction'': ``...'', ``Response 1'': ``...'', ``Response 2'': ``...'', ``Response 3'': ``...'', ``Response 4'': ``...'', ``Response 5'': ``...'' \}
\end{mdframed}
In this case, the prompt includes an instruction and five responses, and ChatGPT is asked to rank these responses at once. 

\section{Experiments}

\begin{table*}[t!]
\caption{Main results. }
\begin{center}
\begin{tabular}{c|cc|cc} 
\hline 
 \multirow{2}{*}{} & \multicolumn{2}{c|}{\textbf{Spearman correlation coefficient}} & \multicolumn{2}{c}{\textbf{Top 1 consistency rate}} \\
 \cline{2-5}
 & \textbf{one-by-one} & \textbf{one-for-all} & \textbf{one-by-one} & \textbf{one-for-all} \\
\hline
NLP tasks & 0.3030 & \textbf{0.3528} & \textbf{0.6050} & 0.4533  \\
\hline
Brainstorming & \textbf{0.6267} & 0.6135 & \textbf{0.8216} & 0.3797  \\
\hline
Generation & \textbf{0.8014} & 0.6244 & \textbf{0.7454} & 0.2611  \\
\hline
OpenQA & \textbf{0.4339} & 0.3781 & \textbf{0.5547} & 0.2688  \\
\hline
Math & 0.3493 & \textbf{0.5134} & \textbf{0.5943} & 0.4339  \\
\hline
Code & 0.4847 & \textbf{0.5321} & \textbf{0.7044} & 0.3643 \\
\hline
Overall & \textbf{0.5321} & 0.5006 & \textbf{0.6836} & 0.3483 \\
\hline
\end{tabular}
\end{center}
\label{consistency}
\end{table*}

\subsection{Datasets}
In this subsection, we introduce the test prompts in our experiment. 
The primary objective of our study is to evaluate the ability of ChatGPT as a comprehensive content preference annotator. 
So our dataset encompasses a diverse range of NLP tasks, along with open-ended instructions such as brainstorming and open QA, which are common online use cases as described in \cite{TrainingLanguageModels2022}.
Table \ref{data_stat} shows details.

\subsection{Models}
In this paper, we focus on Chinese text. 
Models such as LLAMA \cite{touvron2023llama}, OPT \cite{OPTOpenPretrained2022} and GPT-J \cite{gpt-j} have not been particularly optimized in Chinese, so we choose Bloomz \cite{CrosslingualGeneralizationMultitask2022} series models as one of the baseline models. Their pre-training data include more Chinese text, and a further multi-task fine-tuning on Chinese corpus was conducted. 
The following  models are  used to generate responses to each prompt:

  \textbf{Bloomz-176b-mt} \cite{BLOOM176BParameterOpenAccess2022, CrosslingualGeneralizationMultitask2022}, which was further instruction-finetuned on the xP3mt dataset based on Bloom-176b and acquired general prompt understanding abilities to some degree.
  
  \textbf{Bloomz-7b1-mt} \cite{BLOOM176BParameterOpenAccess2022, CrosslingualGeneralizationMultitask2022}, which is the same as Bloomz-176b-mt but with 7.1 billion parameters.

  \textbf{Text-davinci-003}, which is developed by OpenAI and trained to generate high-quality natural language text in a variety of formats through RLHF \cite{TrainingLanguageModels2022, TrainingHelpfulHarmless2022}. 

  \textbf{ChatGPT}, which is similar to Text-davinci-003 but particularly optimized to generate human-like responses to human prompts.

  \textbf{InstrLLM}, which is our own model. InstrLLM has 7 billion parameters and is finetuned on a collection of diversified prompt-response pairs. 
  
\begin{figure*} [t!]
	\centering
	\includegraphics[scale=0.6]{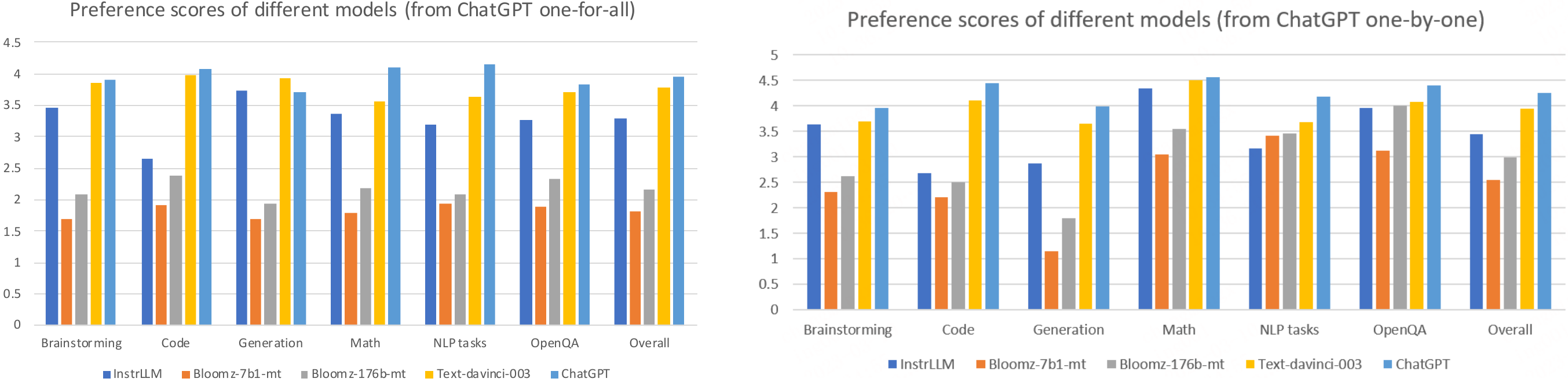}
	\caption{ChatGPT preference scores of different models.}
\label{chatgpt-score}
\end{figure*}

\begin{figure} [t!]
	\centering
	\includegraphics[scale=0.5]{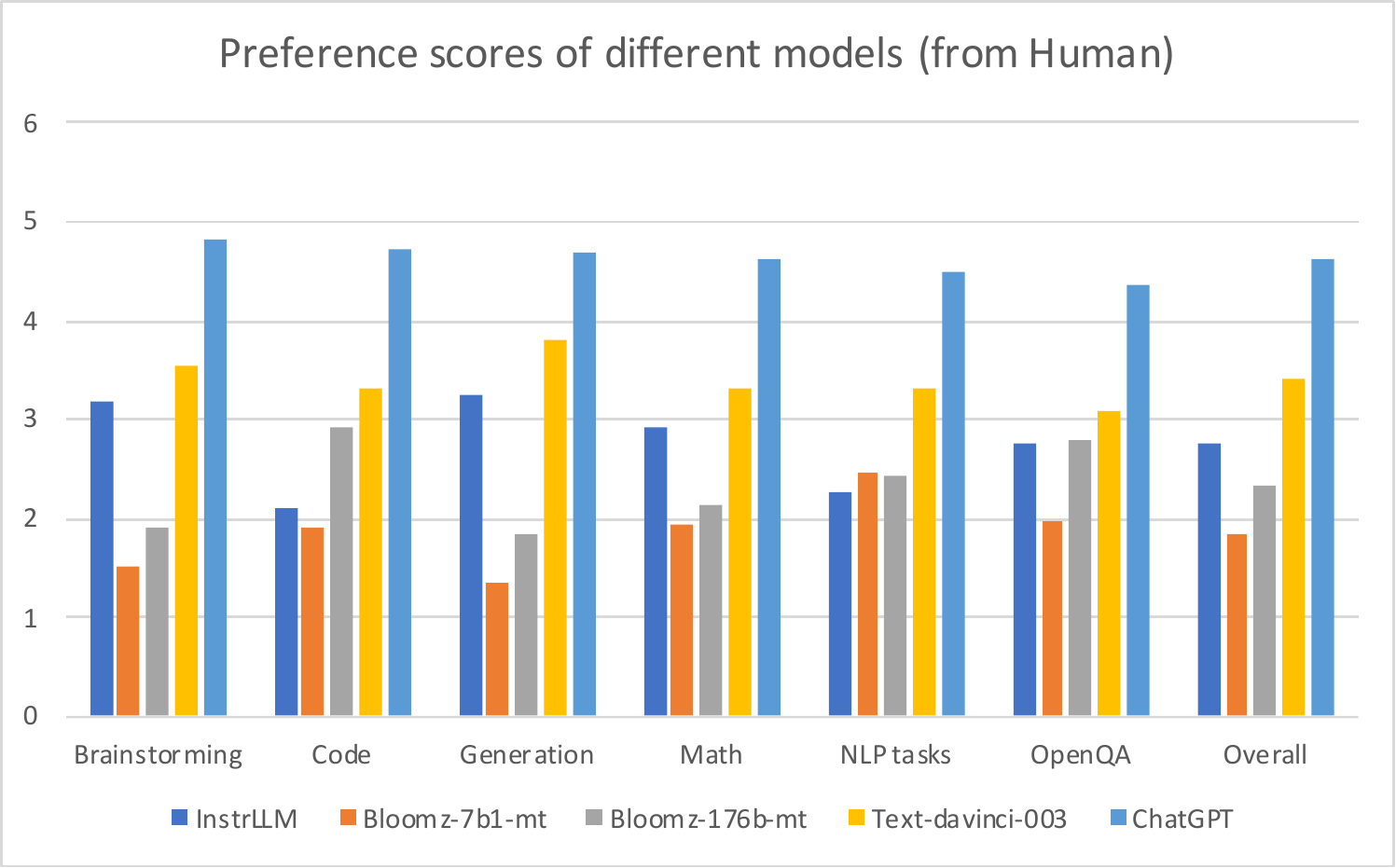}
	\caption{Human preference scores of different models.}
\label{human-score}
\end{figure}

\subsection{Human preference labeling}
To obtain the human preference annotations of the generated responses from  models, we distribute these data for labeling to 129 colleagues, 90\% of whom had a bachelor's degree or higher. 
For each prompt and its corresponding model responses, three annotators independently rank them, and the final ranking is the average of the three. 
We have 30 experienced software developers annotate code-related prompts.

\subsection{Metric}
Spearman correlation coefficient is used to evaluate the consistency between human preferences and ChatGPT preferences among  models' responses.
Additionally, we calculate the top 1 human preference consistency rate to assess the consistency of the first rank model bewteen human and ChatGPT.

\subsection{Main results}
According to Table \ref{consistency}, ChatGPT shows a impressive level of consistency with human preference.  
Overall one-by-one prompting is more consistent with human than one-for-all prompting, especially on brainstorming, generation, and open QA prompts, this is mainly because responses are truncated due to the length limitation while perform one-for-all prompting. 
Meanwhile, one-by-one prompting outperforms one-for-all prompting at all types prompts according to top 1 consistency rate. 
ChatGPT is more consistent with human on brainstorming and generation prompts, and there is still room for improvement for other use cases.
There are two main reasons.
The first one is that these prompts are all relatively difficult for the five  models, so the quality of the generated responses is relatively low, making it difficult to determine which one is better. 
The second one is that to evaluate the responses of these prompts, ChatGPT needs to have strong knowledge and reasoning capabilities.


\subsection{Model Comparison}
A comparison of five models is conducted in six different use cases, as depicted in Figure \ref{human-score} and \ref{chatgpt-score}.
From both Figure \ref{human-score} and \ref{chatgpt-score}, we can see that among all models, ChatGPT demonstrates the best performance, with scores exceeding four points across all use cases. 
Following behind is Text-davinci-003, the gap between ChatGPT and Text-davinci-003 is large from human's perspective  but small from ChatGPT's perspective. 
It is worth noting that our InstrLLM outperforms bloomz-176b-mt overall, especially in brainstorming, generation, and math. 
InstrLLM is adept at brainstorming and generation tasks, which is may because there is a large number of open-ended prompts in its training dataset. 
In Figure \ref{human-score}, Our model performs worst on NLP tasks, which is primarily due to the overfitting to following human instructions.  
Interestingly, from ChatGPT's perspective, our model is comparable with Text-davinci-003, but bloomz-176b-mt, bloomz-7b1-mt are still much worse than Text-davinci-003. 
This indicates that ChatGPT is still not meticulous enough when evaluating contents, and cannot pay attention to many details, such as factualness and ethics. 
When it comes to Code and Math-related prompts, ChatGPT may not be able to check the correctness of code execution and mathematical calculation.

\section{Discussion and Conclusion}
Given the consistency of ChatGPT with human preferences, we discuss the possibilities of using ChatGPT to help us train models, especially in RLHF.
\begin{itemize}
\item 
ChatGPT can be used to construct a set of initial preference annotations to bootstrap reward models.
\item
ChatGPT can also be considered as a  reward model to evaluate the quality of the responses generated by our SFT model. 
Of course, there is the issue that the SFT model trained this way may never be able to surpass ChatGPT, as it is constantly fitting to the preference of ChatGPT.
\item
Based on the zero-shot annotation capability of ChatGPT, we can find out in which use cases the existing  models perform poorly, so that we can further optimize our models for these use cases particularly
\end{itemize}

In this paper, we explore the ability of ChatGPT in content ranking. 
On the diversified test prompts we built, ChatGPT shows a certain consistency with human preference for model responses ranking. 
Based on this discovery, we discuss the possibilities of leveraging ChatGPT to help us train and optimize our own model.


\section{Acknowledgements}
We want to express our gratitude to our colleagues for their professional and efficient data labeling, which has greatly helped us complete the experiment.

\bibliographystyle{unsrtnat}
\bibliography{mainTemplatePDF}

\label{appda}

\end{document}